# Assessing the AI Readiness of Construction Datasets: A Multidimensional Operational Framework

Xiaoyu Hou[a], Bo Xiao[a,*], Abdolmajid Erfani[b], Ronghua Xu[c], Yu Cai[c]

[a] *Department of Civil, Environmental, and Geospatial Engineering, Michigan Technological University, Houghton, MI. 49931*

[b] *Department of Civil and Environmental Engineering, University of Washington, Seattle, WA. 98195*

[c] *Department of Applied Computing, Michigan Technological University, Houghton, MI. 49931*

**Abstract:**

The rapid growth of artificial intelligence (AI) applications in construction has increased the demand for datasets suitable for reliable model development, evaluation, and reuse. However, construction datasets vary widely in documentation, quality, governance, and evaluation support, and consistent criteria for assessing AI readiness remain limited. This study develops a Construction Dataset AI Readiness Framework with 30 attributes across four dimensions: Metadata Completeness, Data Quality and Representativeness, Data Governance and Accessibility, and AI Utility. The framework uses attribute-specific scoring, applicability screening, and automated and manual verification. It was evaluated using 21 publicly available datasets across seven data-type groups. Overall readiness scores ranged from 0.720 to 0.964, with a mean of 0.879. Hybrid verification achieved 90.5% agreement between automated suggestions and manually verified scores. The framework supports consistent assessment of heterogeneous construction datasets and identification of readiness gaps.

## 1. Introduction

Artificial intelligence (AI) has been widely applied in construction to support safety management, progress monitoring, quality inspection, infrastructure assessment, productivity analysis, and other project activities [1, 2]. These applications depend on datasets for model training, validation, evaluation, and reuse. With the growing use of computer vision, sensing, building information modeling (BIM), natural language processing, and other data-driven methods, construction datasets have expanded from image and video collections to point clouds, BIM/ Industry Foundation Classes (IFC) models, sensor records, and textual data [2, 3]. Public datasets also support the reproduction of previous studies, comparison of algorithms, and development of new AI applications. The AI readiness of construction datasets directly affects their suitability for model development, evaluation, benchmarking, and reuse.

The availability of a dataset does not necessarily indicate that it is ready for AI use. Existing construction datasets vary considerably in data quality, metadata, annotation, access conditions, data scale, and application coverage [3]. Xiong et al. [3] reviewed 51 publicly available visual construction datasets and reported substantial differences in data fundamentals, modalities, annotation frameworks, and application contexts. Similar challenges also arise in BIM/IFC models, sensor and time-series data, and textual records, where differences in data structure, documentation, access conditions, and preparation can affect subsequent analysis and reuse [2, 4, 5]. Such variation makes it difficult to determine whether a dataset is suitable for model development, evaluation, benchmarking, or transfer to other applications using consistent and reproducible criteria.

Existing data standards and AI-readiness frameworks provide useful foundations for evaluating these issues. General data principles address data quality, metadata, provenance, accessibility, licensing, interoperability, and reuse, while recent AI-oriented studies extend these considerations to annotation, data distribution, task suitability, and compatibility with machine-learning workflows [6-8]. Domain-specific frameworks further indicate that readiness criteria may need to reflect the characteristics of the target field [9]. Construction datasets introduce additional considerations related to project type, construction method, lifecycle phase, application context, and data-collection context [2, 3]. Recent research has also begun to systematize construction dataset development, including the framework proposed by Ma et al. [10] for construction-domain-general large language models. Existing studies, however, mainly address dataset development, individual data modalities, or specific AI applications. A systematic and operational approach for evaluating heterogeneous construction datasets using consistent criteria, explicit applicability rules, reproducible scoring procedures, and scalable verification mechanisms remains lacking.

This study investigates how the AI readiness of construction datasets can be systematically evaluated. Relevant data standards and principles, construction AI datasets, and existing AI-readiness frameworks are reviewed to identify requirements related to metadata, data quality, governance, AI use, and construction context. A multidimensional framework is developed with explicit attributes, applicability rules, scoring criteria, and verification procedures combining automated and manual assessment. The framework is evaluated using 21 publicly available construction datasets across seven data-type groups to examine its applicability and ability to identify readiness differences across datasets. The proposed framework provides a consistent basis for comparing heterogeneous construction datasets and identifying areas for improving their suitability for AI development, evaluation, and reuse.

## 2. Literature and Standards Foundation

### 2.1 Data standards and principles

Data quality has been extensively studied in information systems and data management. Wang and Strong [11] organized data quality into intrinsic, contextual, representational, and accessibility dimensions, while Strong et al. [12] further emphasized that data usefulness depends on the context of use. This multidimensional view is reflected in ISO/IEC 25012 [13], which defines 15 quality characteristics, including accuracy, completeness, consistency, credibility, accessibility, confidentiality, traceability, understandability, availability, and portability. The standard also distinguishes inherent from system-dependent quality, providing a general basis for evaluating dataset reliability and usability.

Research data management extended these considerations toward data discovery and reuse. Wilkinson et al. [14] proposed the FAIR Guiding Principles, under which data and metadata should be Findable, Accessible, Interoperable, and Reusable. FAIR emphasizes persistent identifiers, rich metadata, standardized access protocols, formal knowledge representation, explicit licenses, provenance, and machine actionability [14]. Because FAIR remains intentionally high-level, its implementation varies across communities [15]. Subsequent studies translated the principles into more measurable forms, including exemplar FAIRness metrics [16] and the FAIR Data Maturity Model [17]. These efforts support more systematic evaluation while retaining flexibility for domain-specific implementation.

Metadata standards provide more explicit structures for dataset identification and description. The DataCite Metadata Schema supports consistent identification, citation, discovery, and linking of research outputs [18]. Version 4.7 includes properties such as Identifier, Creator, Title, Publisher, PublicationYear, Subject, Format, Version, Rights, and Description [18]. Fenner et al. [19] further highlighted persistent identifiers, standardized citation metadata, resolvable landing pages, and machine-readable metadata as key practices for scholarly data repositories.

Provenance records how data are generated, transformed, and maintained. Simmhan et al. [20] identified provenance as important for documenting derivation, processing history, reproducibility, and validation, while the W3C PROV model formalized interoperable representations of entities, activities, and agents involved in data production [21]. Provenance concepts are also incorporated into FAIR, DataCite, and DCAT through information on data origin, related resources, creators, and processing relationships [14, 18, 22]. Such information is particularly important when datasets undergo cleaning, transformation, integration, or repeated publication.

Interoperability and machine-readable exchange are also recurring requirements in data standards. The W3C Data Catalog Vocabulary (DCAT) provides a common RDF-based representation for datasets, distributions, catalogs, and data services [23]. Albertoni et al. [21] showed how DCAT evolved to address dataset discovery, provenance, quality information, data services, and cross-catalog interoperability. The W3C Data Quality Vocabulary (DQV) complements DCAT by representing quality dimensions, metrics, measurements, certificates, and user feedback in machine-readable form [24].

Dataset reuse also depends on governance conditions related to access, ownership, rights, privacy, and responsibility. Khatri and Brown [25] proposed a data-governance framework covering data principles, quality, metadata, access, and lifecycle decisions. Abraham et al. [26] further characterized data governance as the combination of organizational responsibilities, governance mechanisms, and controls used to increase data value while managing associated risks. Similar concerns appear in FAIR, DataCite, and ISO/IEC 25012 through requirements related to licensing, rights, accessibility, and confidentiality [13, 14, 18].

Maintenance and versioning extend these considerations across the dataset lifecycle. DataCite and DCAT provide mechanisms for identifying dataset versions and relationships among releases [23]. González-Cebrián et al. [27] proposed a FAIR-compliant framework for dataset versioning and emphasized its role in tracking changes, supporting reproducibility, and enabling informed reuse. Versioning therefore complements provenance by distinguishing successive states of an evolving dataset [20, 27].

The reviewed literature and standards establish recurring requirements for reliable and reusable datasets. These include data quality and fitness for use [11, 12], FAIRness and machine actionability [14-17], persistent identification and structured metadata [18, 19], provenance and traceability [20, 22], interoperable representation of datasets and quality information [21, 23, 24], governance and usage conditions [25, 26], and version management [23, 27]. These foundations support dataset description, management, evaluation, and reuse, but they do not determine whether a dataset is suitable for a specific AI task or development process. Construction-specific data characteristics and AI-oriented readiness requirements are examined in the following sections.

## 2.2 Construction AI datasets

Publicly available datasets have become an important resource for AI development and evaluation in construction. Reviews of computer vision, wearable sensing, natural language processing, and automated construction monitoring show a growing dependence on domain-specific data across construction management and engineering applications [1, 4, 5, 28]. These studies also indicate that construction AI draws on increasingly diverse data sources, while the availability, scale, and maturity of datasets vary considerably across applications and data types.

Image and video datasets represent the most established category of construction AI data. Xiong et al. [3] recently identified 51 publicly available visual construction datasets published between 2005 and 2024, covering different data modalities, annotation schemes, and monitoring applications. Among representative datasets, Xiao and Kang [29] developed the Alberta Construction Image Dataset (ACID) with 10,000 images of ten types of construction machines for object detection. Duan et al. [30] developed the Site Object Detection Dataset (SODA), which contains approximately 20,000 annotated construction-site images covering workers, materials, machines, and site layouts under varied site conditions. Similar datasets have been developed for worker safety, activity recognition, progress monitoring, and infrastructure inspection [28]. The resulting visual datasets differ substantially in image scale, annotation form, object classes, site coverage, and intended AI tasks.

Three-dimensional point-cloud and spatial datasets support applications involving geometric understanding, semantic segmentation, localization, infrastructure inspection, and digital reconstruction. Trzeciak et al. [31] introduced ConSLAM, which combines repeated terrestrial laser scans with synchronized images, LiDAR measurements, inertial data, and reference trajectories from an active construction site. Qiu et al. [32] developed WHU-Railway3D for railway point-cloud semantic segmentation, covering approximately 30 km of railway environments and multiple semantic classes. Liu et al. [33] proposed a benchmark for as-built BIM reconstruction from real-world point clouds, linking semantically labeled point clouds with corresponding BIM representations. Kellner et al. [34] further developed SemanticBridge using annotated 3D scans of bridges collected under different geographic and scanning conditions. These datasets vary considerably in geometric scale, sensing systems, semantic structure, infrastructure type, and spatial context.

BIM and IFC datasets provide structured construction information in which geometry, object semantics, attributes, and relationships can be represented together. Emunds et al. [35] developed IFCNet as a benchmark dataset for IFC entity classification, providing IFC objects from a broad range of entity classes for machine-learning applications. The GNI BIM Dataset [36] provides a larger collection of anonymized IFC building models spanning different model contents and IFC schemas. Related datasets also connect BIM representations with point clouds and semantic labels for reconstruction and model-enrichment tasks [33]. BIM-oriented datasets therefore present data characteristics that differ from conventional visual datasets, particularly in their schema dependence, object hierarchy, semantic relationships, and model structure.

Sensor and time-series datasets are used to capture construction processes, worker activities, environmental conditions, and human states over time. Tian et al. [37] developed the Construction Motion Data Library as an integrated motion dataset for on-site activity recognition, providing motion data for multiple construction activities. Park et al. [38] released the Scaffolding Worker IMU Time-series dataset, which contains synchronized inertial measurements and keypoint information from workers performing construction-related behaviors. Wearable sensing studies have also used physiological and motion signals such as heart rate, brain activity, visual attention, and inertial measurements for safety and health applications [5]. Compared with static datasets, these resources involve additional temporal characteristics related to sampling, sequence continuity, synchronization, and repeated measurements.

Textual, document-based, and tabular data constitute another growing source for construction AI, particularly with the development of natural language processing and large language models. Shamshiri et al. [4] reviewed 205 studies on text mining and natural language processing in construction, covering information sources such as project documents, accident reports, regulations, contracts, and technical records. Zhong and Goodfellow [39] constructed construction-management corpora from academic literature for domain-specific pre-training of language models and evaluated the resulting models on text-classification and named-entity-recognition tasks. Construction studies have also developed task-oriented datasets from safety records, specifications, codes, and project databases for information extraction, classification, question answering, and retrieval-based applications [4, 39]. These datasets commonly involve different levels of structuring, preprocessing, provenance, and access restrictions.

The current construction AI dataset landscape therefore extends across visual, geometric, model-based, temporal, and textual forms. The datasets differ in data representation, preparation and annotation, application domain, project setting, collection environment, geographic coverage, and temporal or spatial characteristics. Some datasets are developed for narrowly defined tasks or project conditions, while others provide broader benchmark resources across multiple classes, sites, or environments. This heterogeneity forms the domain context for considering whether existing approaches to AI data readiness can be applied consistently to construction datasets. The following section reviews existing AI-readiness frameworks and related practices.

### 2.3 AI readiness frameworks

The growing use of machine learning has expanded the discussion of dataset quality toward the broader concept of AI readiness. Priestley et al. [6] reviewed established data-quality frameworks in relation to machine-learning development and found that the relevance of individual quality characteristics varies across stages of the ML lifecycle. Gong et al. [40] similarly examined dataset-quality issues across data collection, annotation, storage, and application, identifying concerns such as completeness, labeling accuracy, class imbalance, distribution consistency, and format conformity. These studies indicate that the suitability of a dataset for AI depends on both its quality characteristics and the specific purpose for which it is used. Recent standards have begun to formalize this perspective. The ISO/IEC 5259 series extends general data-quality concepts to analytics and machine learning, providing AI-oriented measures and process guidance for data labeling, evaluation, and lifecycle management [41].

Dataset documentation has become another major component of AI readiness. Gebru et al. [7] proposed Datasheets for Datasets to document dataset motivation, composition, collection, preprocessing, intended uses, distribution, and maintenance. Pushkarna et al. [42] developed Data Cards to provide structured summaries of dataset origins, annotation procedures, intended use, evaluation practices, and ethical considerations across the dataset lifecycle. These approaches improve transparency and help downstream users judge whether a dataset is appropriate for a specific AI application. Machine-readable documentation further extends this idea toward direct technical integration. Akhtar et al. [43] introduced Croissant as a common metadata format for ML-ready datasets, supporting dataset discovery, portability, structured interpretation, and interoperability across ML tools and repositories.

Several studies have moved from documentation toward explicit and measurable AI-readiness assessment. Hiniduma et al. [44] developed the AI Data Readiness Inspector (AIDRIN), which combines conventional quality measures such as completeness and duplication with AI-specific indicators including class imbalance, feature relationships, fairness, privacy, and FAIR compliance. Majithia et al. recently proposed an actionable framework for AI-ready data that combines technical suitability, data quality, legal compliance, responsible collection, documentation, and lifecycle practices.

Their work emphasizes practical steps that data publishers can take to improve dataset readiness before downstream AI use.

Domain-specific studies further show that AI readiness is influenced by the context in which data are collected and used. Schwabe et al. [9] developed the METRIC framework through a systematic review of data-quality literature for medical AI and identified 15 dimensions for evaluating training and test datasets. The framework evaluates dataset suitability in relation to the intended ML application and emphasizes characteristics including representativeness, informativeness, consistency, and measurement processes. Domagalski et al. [45] addressed AI readiness in clinical research data through a structured data-preparation methodology involving aggregation, missing-data handling, feature transformation, and compatibility with ML algorithms. Their study demonstrated that preparing data for AI requires domain-sensitive processing decisions that preserve the meaning and integrity of the original data. Similar concerns have been reported in other scientific domains. Poduval et al. [46] discussed AI-ready data in space science and highlighted the importance of access, standardized metadata, data formats, labeling, preprocessing, uncertainty, and documentation of instrumental and operational conditions.

Biomedical research has also begun to develop broader readiness criteria that extend beyond conventional data quality. The NIH Bridge2AI Standards Working Group proposed seven dimensions for biomedical dataset AI readiness, including FAIRness, provenance, characterization, ethics, pre-model explainability, sustainability, and computability [47]. The proposed criteria were applied across multiple biomedical datasets and translated into machine-readable metadata, illustrating how general principles can be extended with requirements arising from a specific scientific domain. Although this work is currently available as a preprint, it provides a useful example of domain-oriented AI-readiness development and formal dataset-level evaluation.

Existing AI-readiness research therefore covers several complementary aspects of dataset preparation and evaluation. ML data-quality studies emphasize annotation, balance, distribution, and lifecycle-dependent fitness for use [6, 40, 41]. Documentation frameworks address dataset origin, processing, intended use, maintenance, and transparency [7, 42], while a study extends dataset description toward machine-readable ML interoperability [43]. AIDRIN and more recent actionable frameworks introduce measurable and implementation-oriented readiness criteria [8, 44]. Domain-specific studies further demonstrate that readiness requirements may need to account for application context, measurement processes, data characteristics, and domain-specific sources of uncertainty [9, 46, 47]. These developments suggest that AI readiness is a multidimensional and context-dependent property of datasets. However, existing approaches differ in scope, operational detail, supported data types, and application domain, and no comparable framework has yet been established for the heterogeneous datasets used across construction AI.

### 2.4 Synthesis and research gap

The literature reviewed in the preceding sections shows that dataset readiness is supported by several established but partly separate knowledge bases. General data-quality and stewardship frameworks provide requirements related to accuracy, completeness, metadata, provenance, accessibility, licensing, interoperability, and reuse [11, 13, 14, 18]. AI-oriented studies extend these foundations to characteristics that directly affect model development, including annotation quality, class balance, data distribution, task suitability, evaluation, and machine-readable integration with ML workflows [43, 44]. Domain-specific studies further show that these requirements need to be interpreted within the context in which data are generated and used. The METRIC framework, for example, synthesized general data-quality literature into dimensions tailored to medical AI training and test data [8]. Bridge2AI similarly extended FAIR-based principles with provenance, data characterization, ethics, sustainability, and computability requirements for biomedical datasets [47]. These studies indicate that AI readiness is a multidimensional property whose relevant dimensions and evaluation criteria may vary across application domains. A comparable domain-specific framework has not yet been established for construction datasets.

Construction presents a particularly heterogeneous data environment. Existing datasets include images, videos, point clouds, BIM/IFC models, sensor sequences, and textual records, with substantial variation in annotation, scale, collection setting, and downstream application [3]. Construction AI applications also span different lifecycle phases, including planning, design, construction, operation and maintenance, with uneven data availability and AI use across these stages [2]. Individual datasets may further represent different project types, construction processes, geographic environments, sensing configurations, and site conditions. SODA, for example, was deliberately collected across multiple construction sites, weather conditions, construction phases, and viewpoints, illustrating the importance of contextual variation in construction data [30]. Research on off-site and modular construction further shows that distinct construction methods generate different processes, information structures, and AI application needs [48]. These characteristics suggest that the relevance and representativeness of construction data cannot be fully interpreted without information about the project and collection context.

The requirements for construction dataset AI readiness can therefore be traced through three complementary routes: established data-quality and stewardship principles, AI-specific readiness requirements, and construction-specific contextual characteristics. However, these requirements remain distributed across general data standards, AI-readiness frameworks, and construction-domain studies. No existing approach integrates them into a unified and operational framework for evaluating heterogeneous construction datasets. The following section develops a construction-specific AI-readiness framework and defines its dimensions, attributes, evaluation rules, and scoring procedures.

## 3. Framework Development and Evaluation Method

### 3.1 Framework development process

Building on the literature and standards synthesis presented in Section 2, the framework development process translated identified requirements related to data quality, documentation, governance, AI readiness, and construction context into a unified set of evaluable attributes. Three evidence foundations were considered: general data standards and principles, studies of construction AI datasets, and existing AI-readiness frameworks. Requirements were extracted at the concept level because similar concepts were often expressed using different terminology, scopes, or levels of specificity across sources.

The extracted requirements were reviewed through concept harmonization, consolidation, screening, and domain interpretation. Terms referring to similar concepts were standardized, overlapping requirements were merged, and items with limited relevance to construction datasets were excluded. Requirements that were too broad for direct assessment were reformulated into attributes that could be evaluated using observable dataset characteristics or documented evidence. Construction-domain interpretation was introduced when general standards or AI-readiness frameworks did not sufficiently capture the characteristics, contexts, or intended uses of construction data.

The consolidated attributes were organized through three derivation routes. Direct Adoption was used when an established requirement could be applied to construction datasets with little or no modification. AI-Specific Adaptation was used when an existing requirement required reinterpretation for AI-oriented dataset development, reuse, or evaluation. Construction-Specific Extension was used for requirements introduced to capture domain conditions not explicitly represented in general-purpose data or AI-readiness frameworks, including project type, construction method, lifecycle phase, application context, and data-collection context. The three routes clarify whether an attribute originates from established data practice, AI-oriented adaptation, or construction-specific extension. The development process is summarized in Fig. 1. The final framework contains 30 attributes organized into four dimensions: Metadata Completeness, Data Quality and Representativeness, Data Governance and Accessibility, and AI Utility. Section 3.2 describes the framework structure and attribute organization.

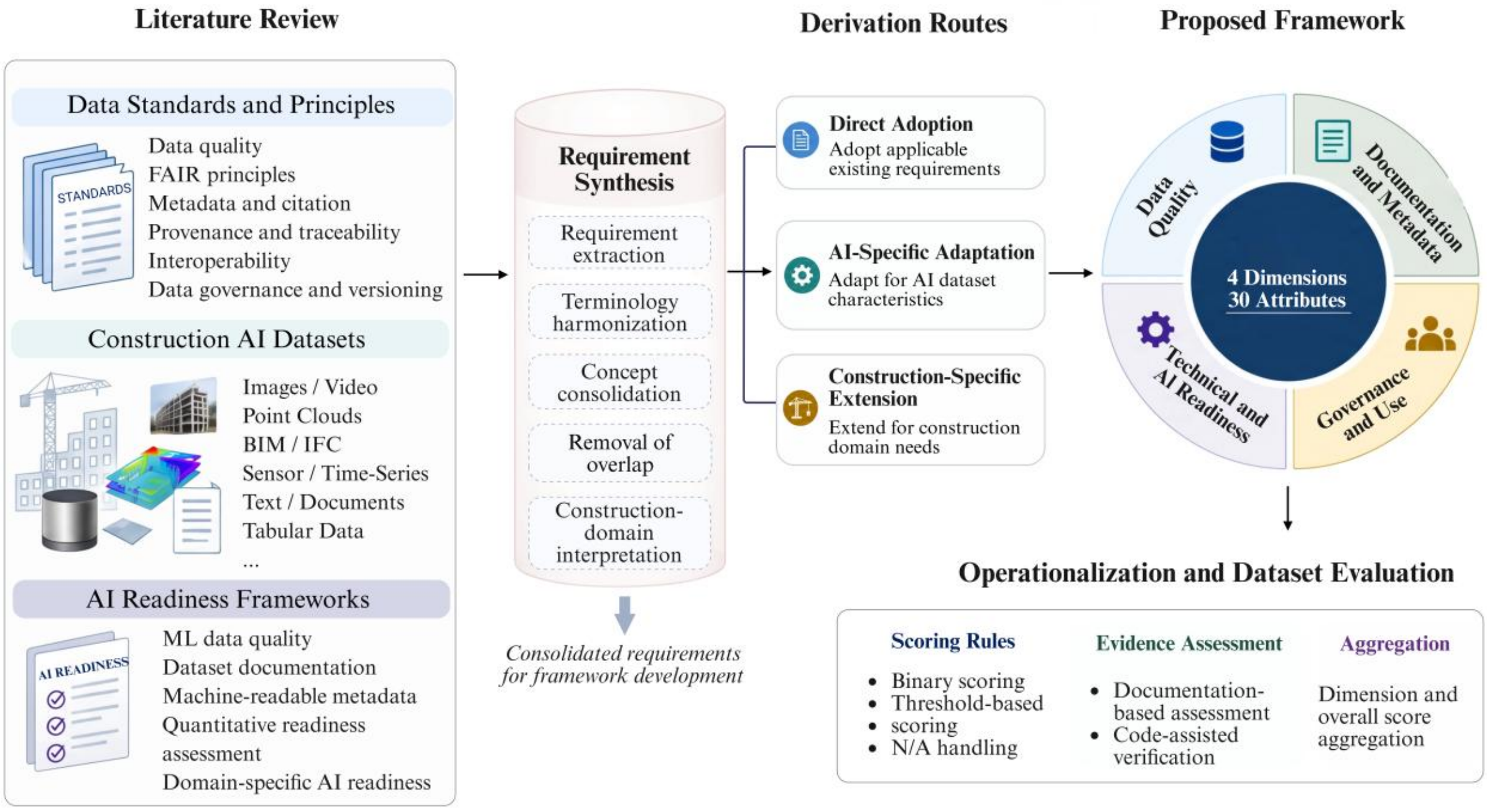

**Fig. 1.** Development process of the Construction Dataset AI Readiness Framework

## 3.2 Framework structure

Each attribute is linked to key supporting sources and one of the three derivation routes defined in Section 3.1. The cited sources identify the main standards, frameworks, or studies used to formulate the attribute. Detailed attribute definitions are provided in the Appendix, while the overall framework structure is presented in Fig. 2.

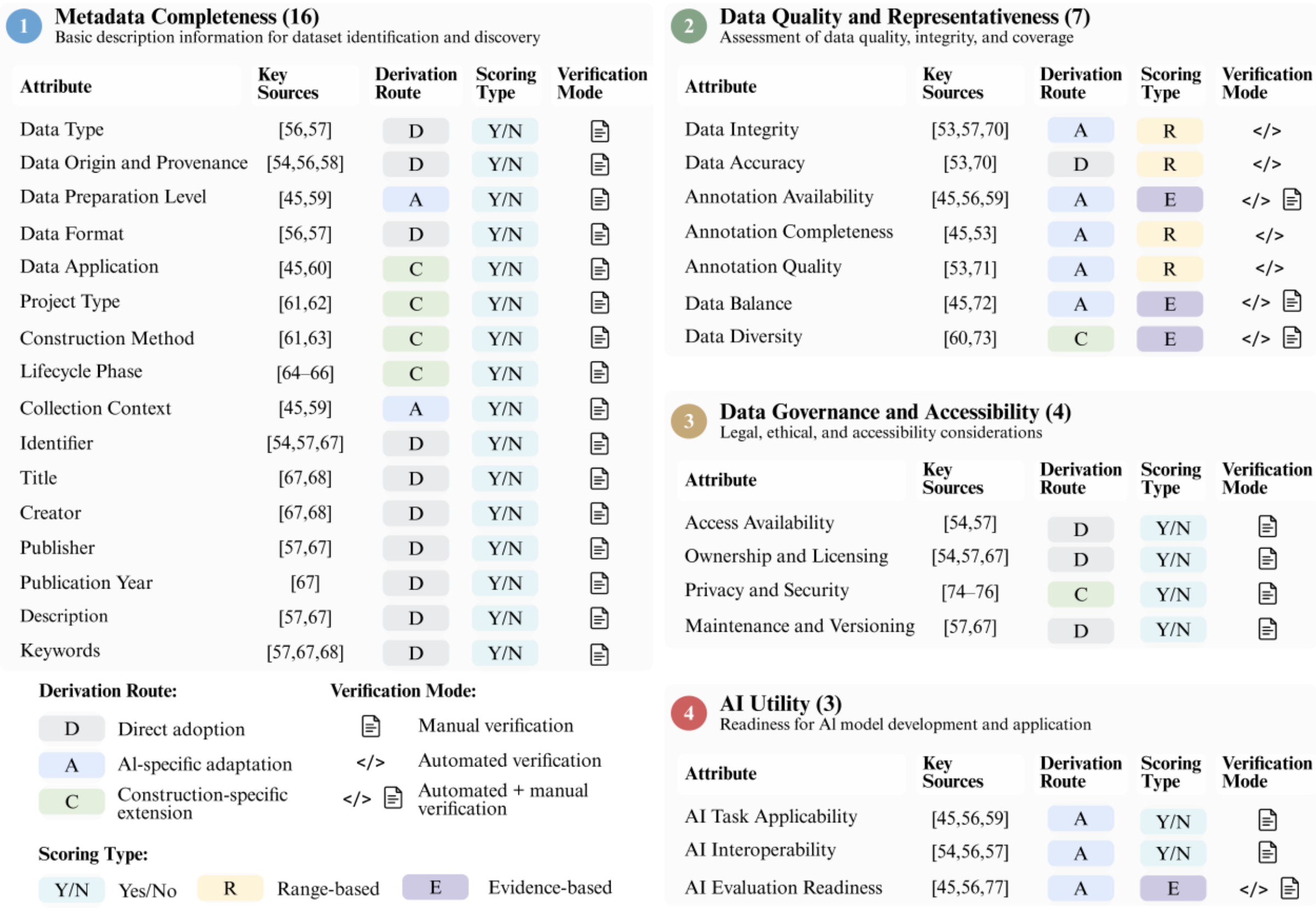

**Fig. 2.** Structure of the construction dataset AI readiness framework [7, 13, 14, 22, 42, 49-67]

**Metadata Completeness** includes 16 attributes describing dataset identity, provenance, preparation, technical form, application context, and descriptive metadata. Established metadata elements include identifier, title, creator, publisher, publication year, description, and keywords. Additional attributes capture construction-related characteristics, including project type, construction method, lifecycle phase, data application, and collection context. These attributes help distinguish datasets that use similar technical formats but represent different project settings, construction processes, or stages of use.

**Data Quality and Representativeness** includes seven attributes: data integrity, data accuracy, annotation availability, annotation completeness, annotation quality, data balance, and data diversity. Data integrity and data accuracy describe the condition and correctness of the underlying data. Annotation availability, completeness, and quality describe the presence and condition of task-relevant labels. Data balance and data diversity address different aspects of dataset composition. Data balance concerns the distribution of samples across task-relevant classes, labels, outcomes, or conditions. Data diversity concerns variation across projects, asset types, locations, environments, operating conditions, and site scenarios.

**Data Governance and Accessibility** includes four attributes: access availability, ownership and licensing, privacy and security, and maintenance and versioning. These attributes describe whether a dataset can be accessed, legally reused, appropriately protected, and maintained over time. They address governance conditions that influence dataset use and continued reuse.

**AI Utility** includes AI Task Applicability, AI Interoperability, and AI Evaluation Readiness. AI Task Applicability assesses whether a dataset supports a defined machine-learning task through identifiable inputs and prediction targets. AI Interoperability assesses machine readability and compatibility with common AI development workflows. AI Evaluation Readiness examines whether the dataset provides resources needed for reproducible model assessment, including predefined data splits, evaluation metrics, baseline results, or evaluation protocols.

Scoring and verification are conducted at the attribute level, as described in Section 3.3.

## 3.3 Evaluation method

Each attribute is evaluated using an evaluation method and a verification mode. The evaluation method defines how the attribute is judged and scored, while the verification mode defines how the supporting evidence is obtained. Three evaluation methods are used: Yes/No, Range-Based, and Evidence-Based Evaluation. Evidence is obtained through Manual Verification, Automated Verification, or Automated + Manual Verification. The combination of these two components allows the same framework to be applied across heterogeneous construction datasets while retaining attribute-specific evaluation procedures.

### 3.3.1 Evaluation methods and verification modes

Yes/No Evaluation is used for attributes assessed from the presence, absence, or documented fulfillment of a defined requirement. Range-Based Evaluation is used when an attribute can be quantified through a reproducible measure $R_i$ derived from dataset files, records, annotations, or other measurable resources. Evidence-Based Evaluation is used when the assessment depends on the sufficiency of available evidence and cannot be determined from a single quantitative measure.

Three verification modes are assigned according to the type of evidence required for each attribute. Manual Verification is used when evaluation depends mainly on dataset documentation, repository information, publications, metadata, or construction-specific interpretation. Automated Verification is used when predefined computational checks can directly produce the measure required for scoring. Automated + Manual Verification is used when computational analysis can identify or summarize relevant evidence, but manual interpretation is still needed to determine the final evaluation outcome.

Among the 30 attributes, 22 use Manual Verification, four use Automated Verification, and four use Automated + Manual Verification. The four fully automated attributes are Data Integrity, Data Accuracy, Annotation Completeness, and Annotation Quality. Annotation Availability, Data Balance, Data Diversity, and AI Evaluation Readiness use Automated + Manual Verification. The evaluation structure is summarized in Fig. 3.

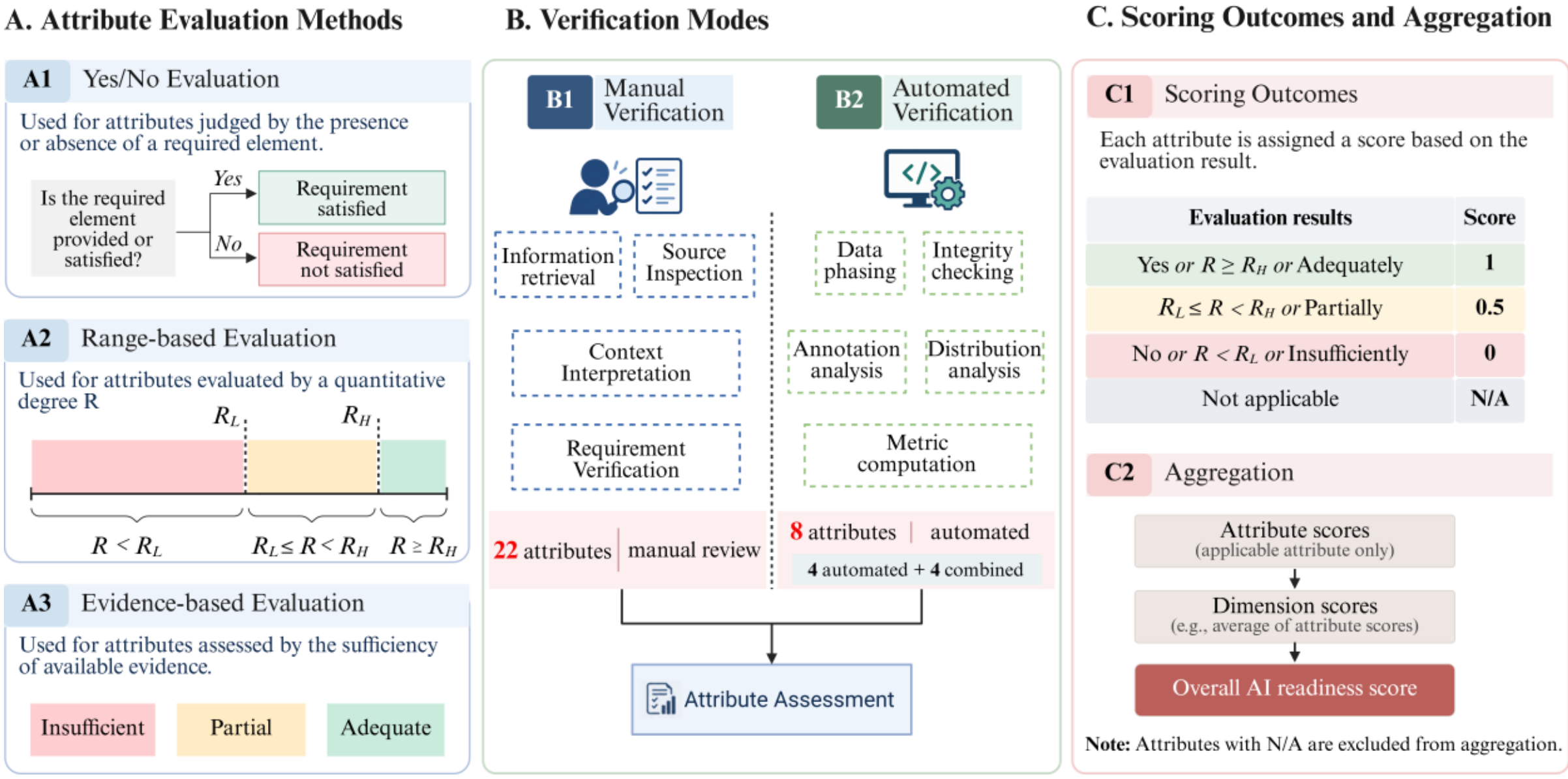

**Fig. 3.** Evaluation method for the Construction Dataset AI Readiness Framework

### 3.3.2 Automated verification

Automated verification was implemented for eight attributes using a Python-based evaluator. The evaluator reads available dataset files, annotations, metadata, and supporting resources and applies predefined checks according to the attribute and available data structure. Two verification pathways are used, as shown in Fig. 4.

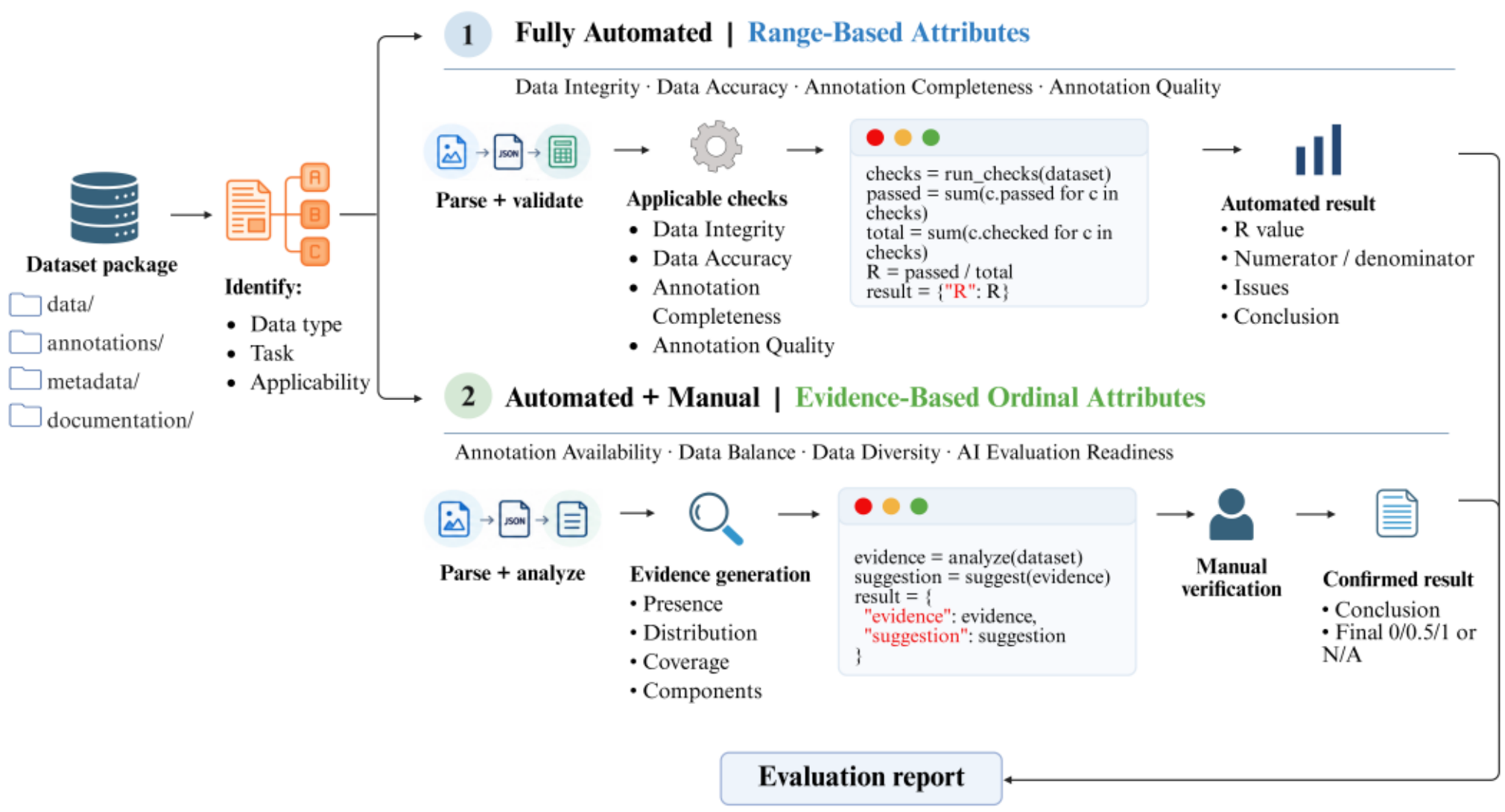

**Fig. 4.** Automated computational verification pathways and outputs

The first pathway is used for Data Integrity, Data Accuracy, Annotation Completeness, and Annotation Quality. These attributes are evaluated through automated quantitative checks. Depending on the attribute, the checks include file readability and consistency, comparison with available reference information, annotation coverage, and structural or geometric validation. For attributes evaluated through unit-level checks, the quantitative measure can be expressed in the general form:

$$R_i = N_{(i,pass)}/N_{(i,checked)} \tag{1}$$

where $N_{i,pass}$ is the number of applicable units satisfying the specified checks and $N_{i,checked}$ is the total number evaluated. The evaluator also records the numerator and denominator, component-level results, detected issues, and a descriptive summary. The resulting $R_i$ is converted to an attribute score using the Range-Based Evaluation criteria described in Section 3.3.3.

The second pathway is used for Annotation Availability, Data Balance, Data Diversity, and AI Evaluation Readiness. The evaluator identifies and summarizes available evidence related to annotation resources, class or condition distributions, measurable contextual variation, and evaluation resources such as data splits, metrics, baseline results, and evaluation protocols. These outputs are reviewed together with dataset documentation, scope, and intended use. The final score is assigned using the Evidence-Based Evaluation criteria described in Section 3.3.3.

Outputs from both pathways are retained at the attribute level. For fully automated attributes, the computational result is used directly for scoring. For hybrid attributes, the computational output serves as evidence for the final manual assessment.

### 3.3.3 Scoring and aggregation

Each attribute is first screened for applicability to the dataset being evaluated. Attributes that do not apply to the dataset are recorded as N/A and excluded from score calculation. Applicable attributes are assigned a score according to their designated evaluation method. All scores are mapped to a common scale from 0 to 1. Table 1 summarizes the scoring criteria for the three evaluation methods. Yes/No Evaluation distinguishes between satisfaction and non-satisfaction of a defined requirement. Range-Based Evaluation maps the quantitative result $R_i$ to three levels using lower and upper thresholds, $T_L$ and $T_H$. Evidence-Based Evaluation uses three levels based on the sufficiency of the available evidence.

**Table 1.** Scoring criteria for the three attribute evaluation methods

| Evaluation method | Score = 1 | Score = 0.5 | Score = 0 |
|---|---|---|---|
| Yes/No | Provided | — | Not provided |
| Range-based | $R_i \geq T_H$ | $T_L \leq R_i < T_H$ | $R_i < T_L$ |
| Evidence-based | Adequate | Partial | Insufficient |

For each dimension $d$, the dimension score $S_d$ is calculated as the mean score of its applicable attributes:

$$S_d = \frac{\sum_{i=1}^{n_d} a_i s_i}{\sum_{i=1}^{n_d} a_i} \tag{2}$$

where $S_i$ denotes the score of attribute $i$, $n_d$ is the number of attributes assigned to dimension $d$, and $a_i$ is an applicability indicator, with $a_i = 1$ for applicable attributes and $a_i = 0$ for attributes recorded as N/A.

The four dimensions contain different numbers of attributes. Their dimension scores are calculated independently and assigned equal weight in the overall AI readiness score:

$$S_{overall} = \frac{1}{4}\sum_{d=1}^{4} S_d \tag{3}$$

Both $S_d$ and $S_{overall}$ range from 0 to 1, with higher values indicating a greater level of AI readiness among the applicable attributes. The number of applicable attributes is retained as part of the evaluation record because datasets with similar readiness scores may be evaluated against different sets of attributes.

### 3.4 Evaluation procedure

Each dataset is evaluated at the attribute level using the definitions established in the framework. For a given attribute, the evaluator reviews the available sources relevant to that requirement, including dataset documentation, repository information, accompanying publications, metadata, and computational outputs where applicable. The assessment is based on the combined evidence available for that attribute.

A score is assigned after the evidence has been reviewed according to the corresponding evaluation method. Along with the score, the evaluator provides a brief rationale summarizing the main evidence and the basis for the judgment. Attributes that do not apply to the dataset are recorded as N/A, while cases in which the available evidence is not sufficient to support a final assessment are flagged for further review.

The completed attribute assessments are used to calculate the dimension-level and overall AI readiness scores according to the aggregation method described in Section 3.3.3. The final evaluation record therefore contains the attribute score, a concise justification, and the supporting evidence used in the assessment.

## 4. Framework Evaluation

### 4.1 Case study design and dataset selection

The case study included 21 publicly available construction datasets covering seven data-type groups: RGB/Image, 3D Point Cloud, BIM/IFC, Sensor/Time-Series, Tabular/Project Data, Document/Text, and Multimodal. Three datasets were selected for each group to provide balanced coverage across the seven categories. The selected datasets represent common forms of construction data and support a range of AI-related tasks, including object detection, semantic segmentation, classification, optimization, knowledge representation, activity recognition, and multimodal analysis. For consistent reporting, the datasets were assigned identifiers from Dataset 1 to Dataset 21. Their names and data-type groups are listed in Table 2.

**Table 2.** Datasets included in the case study

| Dataset ID | Dataset | Data-Type Group |
|---|---|---|
| Dataset 1 | ACID [29] | |
| Dataset 2 | ConstructionSite-10K [68] | RGB / Image |
| Dataset 3 | RDD2022 [69] | |
| Dataset 4 | WHU-Railway3D [32] | |
| Dataset 5 | BIMNet [33] | 3D Point Cloud |
| Dataset 6 | SemanticBridge [34] | |
| Dataset 7 | GNI BIM Dataset [36] | |
| Dataset 8 | IFCNet [35] | BIM / IFC |
| Dataset 9 | Architectural and MEP BIM Object Dataset [70] | |
| Dataset 10 | SWIT Dataset [38] | |
| Dataset 11 | LEAD1.0 [71] | Sensor / Time-Series |
| Dataset 12 | VTT-ConIoT [72] | |
| Dataset 13 | WBLCA Benchmark v2 [73] | Tabular / Project Data |

| Dataset ID | Dataset | Data-Type Group |
| --- | --- | --- |
| Dataset 14 | Construction Time–Cost Trade-off Dataset [74] | |
| Dataset 15 | O-ESD [75] | |
| Dataset 16 | CISOL [76] | |
| Dataset 17 | WikiArch [77] | Document / Text |
| Dataset 18 | BMKG [78] | |
| Dataset 19 | ConSLAM [31] | |
| Dataset 20 | CWPV [79] | Multimodal |
| Dataset 21 | CEQuest [80] | |

Each dataset was independently evaluated by three evaluators using the 30 attributes defined in the framework. The evaluators followed the same attribute definitions, applicability rules, scoring criteria, and evidence requirements. Differences in applicability judgments, attribute scores, and supporting evidence were reviewed jointly after the independent evaluation. Final results were determined through discussion and consensus.

The 21 datasets produced 630 potential dataset assessments before applicability screening. Attributes judged not applicable were recorded as N/A and excluded from score aggregation according to the procedure defined in Section 3. All evaluation evidence was obtained from publicly accessible sources, including released dataset files, metadata records, repository pages, dataset or project websites, accompanying publications, technical documentation, and other supporting materials. The eight attributes designated for code-assisted verification were assessed using the computational procedures defined in Section 3.3.2. The remaining attributes were assessed using their assigned verification modes. Attribute-level results were aggregated into dimension scores and an overall AI readiness score for each dataset.

## 4.2 Evaluation coverage and verification outcomes

### 4.2.1 Overall AI readiness results

Figure 5 shows that AI readiness was generally high across the 21 datasets, but the four dimensions contributed differently to the observed variation. Metadata Completeness was the most consistently strong dimension, with all seven data-type groups clustered near the upper end of the scale. Data Governance and Accessibility showed the greatest separation among groups and the widest within-group dispersion. This contrast suggests that differences in overall readiness were driven less by basic dataset description and more by governance, access, privacy, licensing, and maintenance conditions.

Data Quality and Representativeness showed a more moderate pattern. Most groups remained close to or above the 21-dataset benchmark, while Multimodal datasets had the lowest group mean at 0.794. AI Utility was also strong overall, although its distribution was less uniform. RGB/Image, 3D Point Cloud, and Document/Text reached a group mean of 1.000, whereas Multimodal datasets averaged 0.778. The dimension-level results therefore indicate that strong technical usability did not always coincide with equally strong governance or representativeness.

The overall AI readiness scores ranged from 0.720 to 0.964, with an all-dataset mean of 0.879. Document/Text datasets achieved the highest group mean at 0.939, followed by Tabular/Project Data at 0.902 and RGB/Image at 0.896. Multimodal, 3D Point Cloud, and Sensor/Time-Series groups had lower averages, but their distributions overlapped substantially with those of the higher-scoring groups.

The results do not suggest a simple ranking by data type; readiness remained dataset-specific within each category.

Within-type variability further illustrates this point. Sensor/Time-Series showed the largest dispersion in overall readiness (SD = 0.124), while Document/Text (SD = 0.033) and Tabular/Project Data (SD = 0.037) were comparatively consistent. Multimodal datasets also showed notable spread. These differences indicate that data type alone does not determine readiness, and datasets with similar technical forms can differ considerably in quality, governance, and evaluation support.

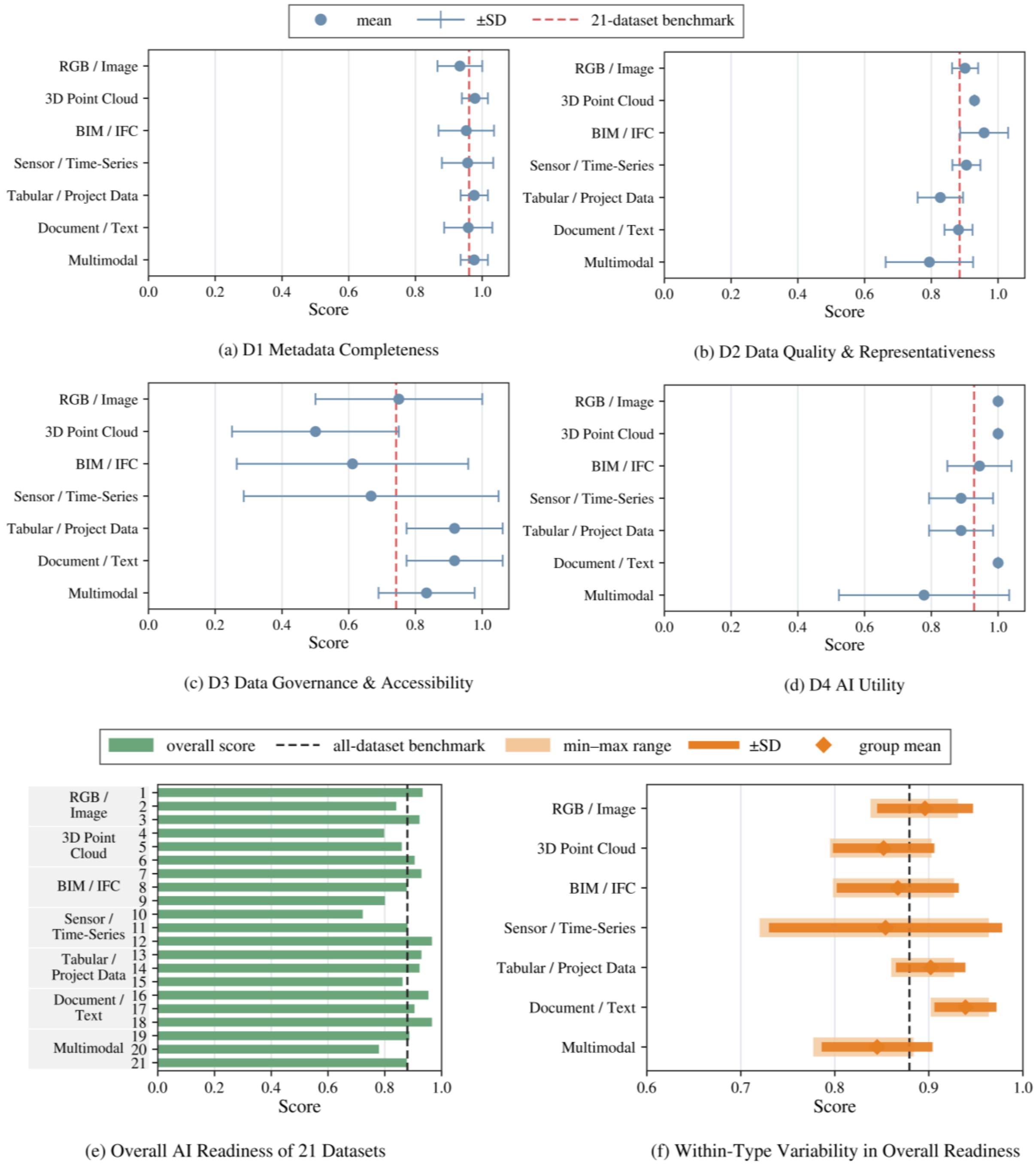

(a) D1 Metadata Completeness (b) D2 Data Quality & Representativeness

(c) D3 Data Governance & Accessibility (d) D4 AI Utility

(e) Overall AI Readiness of 21 Datasets (f) Within-Type Variability in Overall Readiness

**Fig. 5.** Overall AI readiness results

4.2.2 Attribute-level distribution and discrimination

Figure 6 shows that 18 of the 30 attributes (60.0%) varied across the 21 datasets, while the remaining 12 were unchanged. The variation was concentrated in attributes that capture contextual relevance, representativeness, governance, and evaluation support, which are also the areas where heterogeneous construction datasets differ most in practice.

Score variation was especially pronounced for Data Balance, Data Diversity, Privacy and Security, Maintenance and Versioning, and AI Evaluation Readiness. Data Balance was dominated by partial scores, with 81.0% of datasets rated 0.5, indicating that distributional information was often available but rarely sufficient to demonstrate full representativeness. Data Diversity showed a similar pattern at a lower level, with 42.9% receiving 0.5. Privacy and Security produced the clearest deficiency signal, as 52.4% of datasets received a score of 0. These patterns suggest that differences among datasets were driven less by basic technical usability and more by the completeness of evidence needed to support broader reuse, governance, and evaluation.

Applicability variation formed a second source of discrimination. Construction Method was recorded as N/A for 76.2% of the datasets, largely because many public datasets describe assets, scenes, sensor records, or knowledge structures without a direct link to a defined construction production process. Project Type and Lifecycle Phase showed similar, although less pronounced, applicability differences. Annotation-related attributes also became N/A when a dataset did not contain a distinct annotation or label layer. These cases show that some construction-specific and task-specific attributes are meaningful only under particular dataset structures and use contexts.

The most informative attributes were those that produced three distinct outcomes across the sample. Project Type, Construction Method, Data Accuracy, Annotation Availability, Annotation Quality, and Privacy and Security each combined score differences with applicability differences. Their distributions therefore captured more than a single readiness level and provided stronger separation among heterogeneous datasets. In contrast, many metadata and technical attributes remained concentrated at score 1, indicating that these requirements were commonly satisfied across the selected public datasets and contributed less to cross-dataset differentiation.

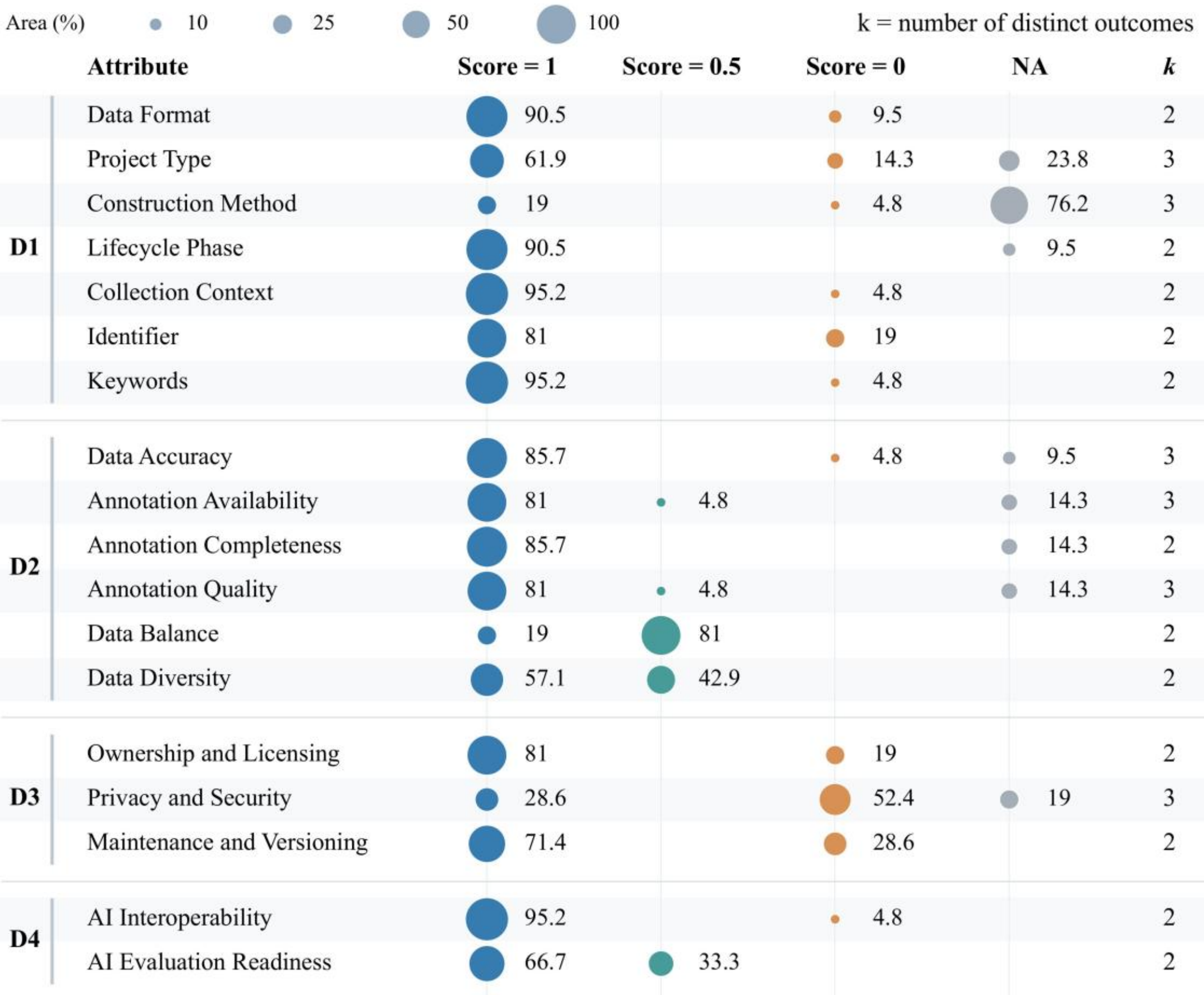

**Fig. 6.** Attribute-level score distribution for the 18 variable attributes across the 21 datasets

#### 4.2.3 Automation–manual agreement

The four hybrid attributes generated 84 suggested scores across the 21 datasets. Manual verification confirmed 76 of these scores, corresponding to an overall agreement rate of 90.5%. The remaining eight cases required revision, with six upward corrections from 0.5 to 1 and two downward corrections from 1 to 0.5.

Agreement was consistently high for Annotation Availability, Data Balance, and Data Diversity, each reaching 95.2%. AI Evaluation Readiness showed greater divergence, with an agreement rate of 76.2% and five revised scores. This pattern indicates that the automated evidence was generally sufficient for the first three attributes, while evaluation readiness more often required additional interpretation of documentation, benchmarks, metrics, or protocols that were not fully captured by the automated evidence alone. The direction of the corrections also provides useful context. Most disagreements were upward revisions, showing that the automated procedure more often underestimated the available evidence than overstated it. Downward corrections were limited to two cases. The hybrid assessment therefore showed strong consistency, while retaining a clear role for manual review in attributes that depend more heavily on contextual evidence.

**Table 3.** Agreement between suggested and manually verified scores for the four hybrid attributes

| Attribute | Total Comparisons | Agreements | Disagreements | Agreement Rate |
|---|---|---|---|---|
| Annotation Availability | 21 | 20 | 1 | 95.2% |
| Data Balance | 21 | 20 | 1 | 95.2% |
| Data Diversity | 21 | 20 | 1 | 95.2% |

| Attribute | Total Comparisons | Agreements | Disagreements | Agreement Rate |
|---|---|---|---|---|
| AI Evaluation Readiness | 21 | 16 | 5 | 76.2% |
| **Overall** | **84** | **76** | **8 (↑6, ↓2)** | **90.5%** |

Note： ↑ indicates an upward manual correction (0.5 → 1); ↓ indicates a downward manual correction (1 → 0.5).

#### 4.2.4 N/A and applicability patterns

Figure 7 shows that N/A outcomes accounted for 38 of the 630 assessments (6.0%), with most cases concentrated in a small subset of context-dependent attributes. Construction Method contributed 16 N/A cases, followed by Project Type with five and Privacy and Security with four. Together, these three attributes accounted for 65.8% of all N/A outcomes. This distribution indicates that applicability was shaped mainly by the extent to which a dataset could be linked to a defined construction context, annotation structure, or governance requirement.

This concentration is also visible at the dimension level. D1 and D2 together accounted for 34 of the 38 N/A outcomes (89.5%), reflecting the stronger dependence of metadata context and annotation-related quality attributes on dataset structure and intended use. D4 contained no N/A cases, meaning that AI Task Applicability, AI Interoperability, and AI Evaluation Readiness could be assessed for every dataset in the sample.

Data type further shaped where applicability issues occurred. Tabular/Project Data had the highest N/A rate (11.1%), followed by BIM/IFC (8.9%) and Multimodal data (7.8%); these three groups contributed 25 of the 38 N/A cases. Combined with the high N/A frequency of Construction Method, these results show that applicability differences were concentrated in specific attributes and data-type groups. This pattern shows that a common 30-attribute framework can retain broad coverage while allowing specific requirements to apply selectively across heterogeneous construction data.

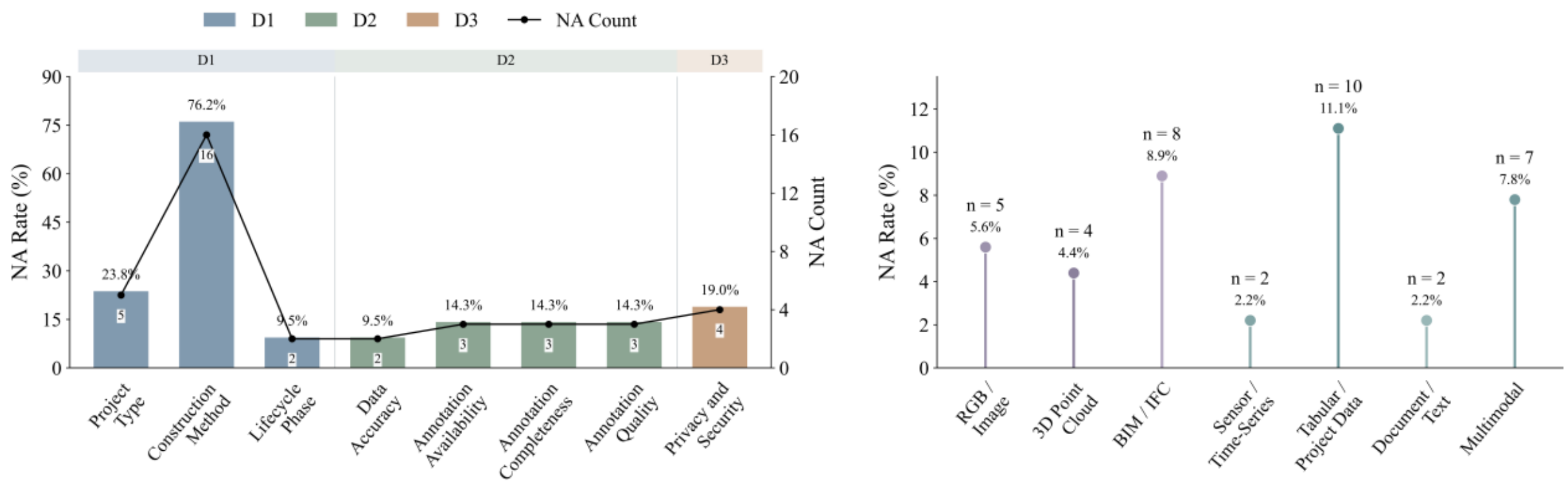

(a) Attribute-level NA occurrence (b) NA occurrence across data type groups

**Fig. 7.** Distribution of N/A outcomes across attributes and data-type groups.

## 5. Discussion

### 5.1 Sensitivity of automated scores to threshold selection

Because the range-based attributes are converted from continuous $R$ values into ordinal readiness scores, the selected thresholds can influence cases located near a decision boundary. The sensitivity analysis therefore examined whether reasonable changes in the threshold settings would alter the automated scoring results. For this study, $T_H = 0.90$ was selected as the baseline requirement for full readiness, requiring at least 90% of the applicable automated checks to be satisfied. $T_L$ = 0.70 was used to distinguish partial readiness from insufficient performance. To examine the effect of these choices, the high threshold varied between 0.90 and 0.95 and the low threshold between 0.60 and 0.70, producing four scenarios: S1 (0.90/0.60), S2 (0.90/0.70), S3 (0.95/0.60), and S4 (0.95/0.70). S2 was used as the baseline (Table 4).

**Table 4.** Sensitivity of automated evaluation scores to alternative threshold settings

| Scenario | $T_H$ | $T_L$ | Score = 1 | Score = 0.5 | Score = 0 | Mean Score | Changes vs. S2 |
|---|---|---|---|---|---|---|---|
| S1 | 0.90 | 0.60 | 74 (97.4%) | 1 (1.3%) | 1 (1.3%) | 0.980 | 0 |
| S2 (Baseline) | 0.90 | 0.70 | 74 (97.4%) | 1 (1.3%) | 1 (1.3%) | 0.980 | — |
| S3 | 0.95 | 0.60 | 73 (96.1%) | 2 (2.6%) | 1 (1.3%) | 0.974 | 1 (1.3%) |
| S4 | 0.95 | 0.70 | 73 (96.1%) | 2 (2.6%) | 1 (1.3%) | 0.974 | 1 (1.3%) |

The automated scores were stable across the tested threshold settings. Of the 76 evaluations, 75 (98.7%) retained the same ordinal score under all four scenarios. Changing $T_L$ from 0.60 to 0.70 produced no score changes. The only sensitivity occurred when $T_H$ increased from 0.90 to 0.95. Dataset 11 had a Data Integrity value of $R = 0.943$, which received a score of 1 under S1 and S2 and 0.5 under S3 and S4. The mean automated score consequently decreased only from 0.980 to 0.974.

The limited influence of $T_L$ reflects the observed distribution of $R$ values. None of the applicable measurements fell between 0.60 and 0.70, so shifting the lower cutoff within this interval did not affect the scores. The only sensitive case was close to the upper decision boundary. Values well above 0.95 or below 0.60 retained the same classification, whereas a value between 0.90 and 0.95 could shift from full to partial readiness under the stricter high threshold. This shows that threshold selection matters mainly for cases close to a decision boundary.

These results support the stability of the baseline setting for the present case study. The overall findings did not materially depend on the tested threshold combinations, with only 1.3% of applicable automated evaluations changing classification. The analysis does not establish a universally optimal threshold pair, however. The thresholds serve as operational decision boundaries and may require reassessment when the framework is applied to datasets with different quality distributions. Reporting the underlying $R$ values alongside the ordinal scores is therefore useful for cases near a cutoff.

### 5.2 Implications of applicability-adjusted scoring for dataset comparison

The exclusion of N/A attributes makes each readiness score conditional on the attributes applicable to that dataset. Applicability was uneven across the case study, with 38 assessments classified as N/A and most cases concentrated in a small number of context-dependent attributes and data-type groups. As a result, datasets with similar dimensions or overall scores may have been evaluated against different sets of attributes.

This affects cross-dataset interpretation. A higher score indicates stronger performance among the applicable attributes, but it does not indicate broader evaluation coverage. The number and composition of applicable attributes provide complementary information about the basis of the score, particularly for dimensions containing construction-context and annotation-related attributes. Reporting applicability coverage together with readiness scores would make dataset comparisons more interpretable and distinguish score magnitude from evaluation coverage.

### 5.3 Readiness gaps in public construction datasets

The case study shows that basic metadata and technical usability were consistently strong across the evaluated datasets. Greater variation appeared in representativeness, governance, maintenance, and evaluation support. Data Balance, Data Diversity, Privacy and Security, Maintenance and Versioning, and AI Evaluation Readiness were among the attributes with the largest score variation, whereas many descriptive metadata attributes were consistently satisfied.

This pattern suggests that the current readiness gap is concentrated in the evidence needed to support dataset reuse and evaluation. Public datasets commonly provide sufficient information to identify, access, and process the data, but information on sample composition, governance conditions, update status, and evaluation resources is less consistent. The distinction is important because these elements affect whether users can judge the scope, limitations, and evaluation conditions of a dataset before applying it to an AI task.

Variation within the same data-type groups reinforces this finding. Datasets with similar technical forms still differed in governance, representativeness, and evaluation support, and several groups showed substantial within-group dispersion in overall readiness. The observed readiness gaps therefore cannot be attributed to data format or modality alone. They are also shaped by supporting information and evaluation resources released with the dataset.

### 5.4 Limitations and future work

Several limitations remain in the current framework and case study. The 30-attribute structure was derived from literature, standards, AI-readiness concepts, and construction-specific requirements, but its construct coverage has not yet been independently validated. The case study showed that 18 attributes varied across the evaluated datasets, while 12 remained unchanged. This result does not establish that the invariant attributes are redundant, but it leaves open questions about attribute completeness, overlap, granularity, and discriminative value. Future work should examine the attribute set through expert validation and larger-sample testing, with refinement where consistent redundancy or missing concepts are identified.

The empirical evaluation was limited to 21 publicly available datasets, with three datasets in each of seven data-type groups. This sampling design provides coverage across major construction data forms but does not capture the full range of industry-owned, continuously updated, or highly specialized datasets. Broader validation should include datasets from additional construction domains, lifecycle stages, and ownership settings to test whether the observed readiness patterns remain stable.

The aggregation scheme also warrants further examination. Dimension scores are calculated from applicable attributes, and the four dimensions contribute equally to the overall readiness score. Differences in applicability mean that similar scores can be based on different attribute sets, while equal weighting assumes comparable importance across dimensions. Future studies can test alternative weighting and aggregation strategies and report applicability coverage alongside readiness scores to assess the stability of cross-dataset comparisons.

A final limitation concerns the external validity of the readiness score. The current study evaluates dataset characteristics and supporting evidence, but it does not test whether higher readiness scores are associated with stronger downstream AI outcomes. Future validation should examine relationships with reproducibility, transferability, evaluation consistency, and data-preparation effort. Establishing these links would provide stronger evidence that the framework captures readiness that is meaningful for practical AI use.

## 6. Conclusion

This study developed a Construction Dataset AI Readiness Framework comprising 30 attributes across four dimensions: Metadata Completeness, Data Quality and Representativeness, Data Governance and Accessibility, and AI Utility. The framework operationalizes AI-readiness assessment through attribute-specific scoring, applicability screening, and verification procedures that combine automated and manual evaluation. Eight attributes incorporate computational verification, including four fully automated and four hybrid attributes.

The framework was evaluated using 21 publicly available datasets spanning seven data-type groups. Metadata Completeness was consistently strong, while greater variation occurred in representativeness, governance, maintenance, and evaluation support. Eighteen of the 30 attributes differentiated datasets within the case study. Data Balance, Data Diversity, Privacy and Security, Maintenance and Versioning, and AI Evaluation Readiness showed the largest variation. These results identify governance, representativeness, and reproducible evaluation support as major areas for improving construction dataset readiness.

Applicability screening and code-assisted verification were important to the operational use of the framework. N/A outcomes accounted for 6.0% of all assessments and were concentrated in a limited set of context-dependent attributes. Hybrid verification achieved 90.5% agreement between automated suggestions and manually verified scores, while threshold sensitivity analysis produced limited changes under the tested settings. The evaluation also showed that directly measurable attributes are more suitable for computational assessment, while contextual, governance, and semantic attributes still require manual interpretation. This hybrid structure supports more scalable assessment across heterogeneous construction datasets.

The framework provides a structured basis for comparing existing datasets, identifying readiness gaps, and guiding future dataset development. Further work should extend the evaluation to broader dataset collections, examine the coverage and independence of the 30 attributes, and test alternative aggregation strategies. Future validation should also examine whether higher readiness scores are associated with improved reproducibility, transferability, evaluation consistency, and reduced data-preparation effort.

**Acknowledgment**

This work is supported by the National Science Foundation (NSF), award # 2550033. Any opinions, findings, and conclusions or recommendations expressed in this paper are those of the author(s) and do not necessarily reflect the views of the NSF.

**Data availability statement**

Some or all data, models, or code that support the findings of this study are available from the corresponding author upon reasonable request.

**Appendix Table A1**

| Attribute | Definition |
|---|---|
| Data Type | The primary modality or form in which the dataset represents information, such as images, videos, point clouds, BIM models, text documents, sensor time series, or tabular records. |
| Data Origin and Provenance | The origin of the data and the history of how they were collected, generated, reused, derived, or transformed, such as data captured from active construction sites, generated through simulation, derived from an existing dataset, or processed from raw sensor records. |
| Data Preparation Level | The stage of processing or transformation at which the dataset is provided, such as raw site images, cleaned sensor records, normalized point clouds, processed BIM data, or annotated images. |
| Data Format | The file format, encoding, or machine-readable representation used to store and exchange the dataset, such as JPEG or PNG images, CSV tables, JSON annotations, IFC models, LAS point clouds, or MP4 videos. |
| Data Application | The construction-related purpose or practical application for which the dataset can be used, such as safety monitoring, progress tracking, infrastructure inspection, equipment monitoring, or construction robotics. |
| Project type | The type of construction project, facility, or infrastructure represented by the dataset, such as residential or commercial buildings, bridges, roads, railways, industrial facilities, or utility infrastructure. |
| Construction Method | The construction production or delivery method represented by the dataset, such as conventional on-site construction, prefabrication, modular construction, off-site manufacturing, or additive construction. |
| Lifecycle Phase | The stage of the built asset lifecycle represented by the dataset, such as planning and design, fabrication, construction, operation and maintenance, renovation, or demolition. |
| Collection Context | The physical, environmental, temporal, or operational conditions under which the data were collected or generated, such as indoor or outdoor sites, day or night conditions, different weather conditions, controlled experiments, active jobsites, or varying camera viewpoints. |
| Identifier | A unique and preferably persistent reference that identifies the dataset, such as a DOI, repository accession number, or other persistent identifier. |
| Title | The official name used to identify and distinguish the dataset from other data resources, such as “Advanced Construction Image Dataset.” |
| Creator | The individual, group, or organization primarily responsible for creating or compiling the dataset, such as a research team, university laboratory, government agency, or industry organization. |

| Attribute | Definition |
| --- | --- |
| Publisher | The organization, repository, or service responsible for formally publishing or making the dataset available, such as Zenodo, Figshare, an institutional repository, or a dedicated dataset hub. |
| Publication Year | The year in which the dataset or the referenced release was formally published or made publicly available, such as 2026 for an initial 2026 release. |
| Description | A concise account of the dataset's purpose, scope, content, and major characteristics, such as the construction activities represented, dataset scale, collection setting, and intended uses. |
| Keywords | Terms that characterize the dataset's main topics, data modalities, or applications and support discovery, such as "construction safety," "BIM," "point cloud," or "progress monitoring." |
| Data Integrity | The extent to which dataset files and records remain technically intact, readable, and internally consistent, such as whether images can be opened, JSON files can be parsed, expected records are present, and linked identifiers remain consistent across files. |
| Data Accuracy | The extent to which data values, measurements, metadata, or records satisfy verifiable correctness checks against available internal or external references. |
| Annotation Availability | Whether annotations or labels needed to use the annotated dataset are provided and accessible, such as class labels, bounding boxes, segmentation masks, keypoints, textual captions, timestamps, or numerical targets. |
| Annotation Completeness | The extent to which applicable samples and annotation targets are fully annotated with the required information, such as whether all intended images, objects, classes, coordinates, masks, or timestamps are covered. |
| Annotation Quality | The extent to which annotations satisfy verifiable structural, geometric, semantic-consistency, and quality-control criteria, including valid class assignments, bounding boxes or masks, consistent labeling, cross-format agreement, and documented annotation quality controls. |
| Data Balance | The distribution of samples across task-relevant classes, labels, outcomes, or conditions relative to the dataset's stated purpose, such as machinery classes, safe and unsafe activities, defect types, or clear and occluded observations. |
| Data Diversity | The extent to which the dataset includes documented variation across relevant data sources, projects, asset types, locations, environments, operating conditions, or site scenarios within its stated scope. |
| Access Availability | Whether and how intended users can obtain the dataset, such as through direct download, a public repository, API access, registration, or an approval-based request mechanism. |
| Ownership and Licensing | Information identifying ownership or release authority and specifying the conditions under which the dataset may be accessed, used, modified, attributed, or redistributed, such as CC BY 4.0, research-only use, or noncommercial-use restrictions. |

| Attribute | Definition |
| --- | --- |
| Privacy and Security | The identification and management of privacy, confidentiality, proprietary, or security risks associated with the dataset, such as worker faces or voices, personal identifiers, sensitive project locations, building designs, or company records protected through anonymization or controlled access. |
| Maintenance and Versioning | Information describing how dataset releases are identified, updated, corrected, maintained, superseded, or archived over time, such as version numbers, release dates, changelogs, update schedules, maintenance contacts, or archive notices. |
| AI Task Applicability | The AI or machine-learning task that can be formulated using the dataset and its associated inputs and prediction targets, such as object detection from construction images, defect classification from inspection imagery, progress recognition from video, or equipment-condition prediction from sensor data. |
| AI Interoperability | The extent to which the dataset can be machine-read, interpreted, and integrated into common AI development tools and workflows, such as through standard formats including CSV, JSON, COCO, IFC, or reusable data loaders and conversion scripts. |
| AI Evaluation Readiness | The extent to which the dataset provides the resources needed for systematic and reproducible evaluation of AI models, such as predefined training, validation, and test splits, evaluation metrics, benchmark baselines, and documented evaluation protocols. |